\documentclass[letterpaper, 10 pt, conference]{ieeeconf}  

\IEEEoverridecommandlockouts                              
\title{\LARGE \bf 
Hybrid Iterative Linear Quadratic Estimation: Optimal Estimation for Hybrid Systems}

\author{J. Joe Payne, James Zhu, Nathan J. Kong, and Aaron M. Johnson
    \thanks{This material is based upon work supported by the U.S. National Science Foundation
    under grant \#CMMI-1943900.}%
    \thanks{All authors are with the Department of Mechanical Engineering, Carnegie Mellon University, Pittsburgh, PA, USA, \texttt{jjpayne@andrew.cmu.edu}, \texttt{amj1@andrew.cmu.edu}}%
}
\usepackage{amsmath}
\usepackage{mathtools}
\usepackage{amsthm}
\usepackage{amssymb}
\usepackage{accents}
\usepackage{cite}

\usepackage{algorithm}
\usepackage{algpseudocode}
\usepackage[bookmarks=true]{hyperref}
\hypersetup{%
  breaklinks,
  bookmarksnumbered=true,
  bookmarksopen=true,
  colorlinks=true,
  linkcolor=black,
  urlcolor=black,
  citecolor=black
}

\newcommand{\V}{\mathcal{V}}
\newcommand{\der}{\textbf{$\mathrm{D}$}}

\newcounter{theorem}

\newtheorem{definition}[theorem]{Definition}

\begin{document}
\maketitle
\thispagestyle{empty}
\pagestyle{empty}
\begin{abstract}
In this paper we present Hybrid iterative Linear Quadratic Estimation (HiLQE), an optimization based offline state estimation algorithm for hybrid dynamical systems.
We utilize the saltation matrix, a first order approximation of the variational update through an event driven hybrid transition, to calculate gradient information through hybrid events in the backward pass of an iterative linear quadratic optimization over state estimates.
This enables accurate computation of the value function approximation at each timestep.
Additionally, the forward pass in the iterative algorithm is augmented with hybrid dynamics in the rollout.
A reference extension method is used to account for varying impact times when comparing states for the feedback gain in noise calculation.
The proposed method is demonstrated on an ASLIP hopper system with position measurements.
In comparison to the Salted Kalman Filter (SKF), the algorithm presented here achieves a maximum of 63.55\% reduction in estimation error magnitude over all state dimensions near impact events.
 
\end{abstract}

\section{Introduction}
Contact is essential for many robots as they must physically interact with their environment to accomplish their goals. 
For example, legged robots must repeatedly make and break contact with the ground to move around and perform their tasks, whether that is mapping, surveying, search and rescue, or any other task.
Similarly, manipulation robots must make contact with the objects they seek to manipulate and their environment in order to push, pull, grasp, etc., those objects.
To plan reliable control sequences through contact, robots need an accurate state estimate.
However, intermittent contacts cause robots' dynamics to become non-smooth and potentially discontinuous, breaking the assumptions of classical methods of state estimation which assume smoothness \cite{bloesch2013state,mitcheetahstateestimation2018, varin2018constrained, hartley2020contact, BaillyOptimalEstimation2021}.

Accurate state estimation is especially critical near contact surfaces as minor configuration variations can cause major changes in control authority.
A key example is when a foot is in contact with the ground, that leg can exert large ground reaction forces on the robot, but if that same foot is even slightly off the ground, then it can exert no force.
Accordingly, for controllers to generate feasible plans, a strong estimate of current state is necessary.
This is typically approached as a filtering problem using either Kalman filters \cite{bloesch2013state, hartley2020contact,kong2021salt,GAO2021290,paper:payne-uncertainty-2022} or particle filters \cite{koutsoukos2002monitoring,koval2017manifold}.

There are many situations where full state ground truth cannot be obtained to evaluate the performance of the robot or tuning of these algorithms.
In these cases, offline methods for state estimation \cite{Zhang2020,BaillyOptimalEstimation2021} can provide a log of the robot's behavior as well as a comparison point for estimation performance when tuning gains and designing online estimation algorithms. These offline methods can provide better performance as they consider all of the information from a given trial to achieve higher state estimation accuracy at each timestep.
These frameworks can also be used to better estimate dynamic parameters of a system given the additional information.

\begin{figure}[tb]
    \centering
    \includegraphics[width = \linewidth]{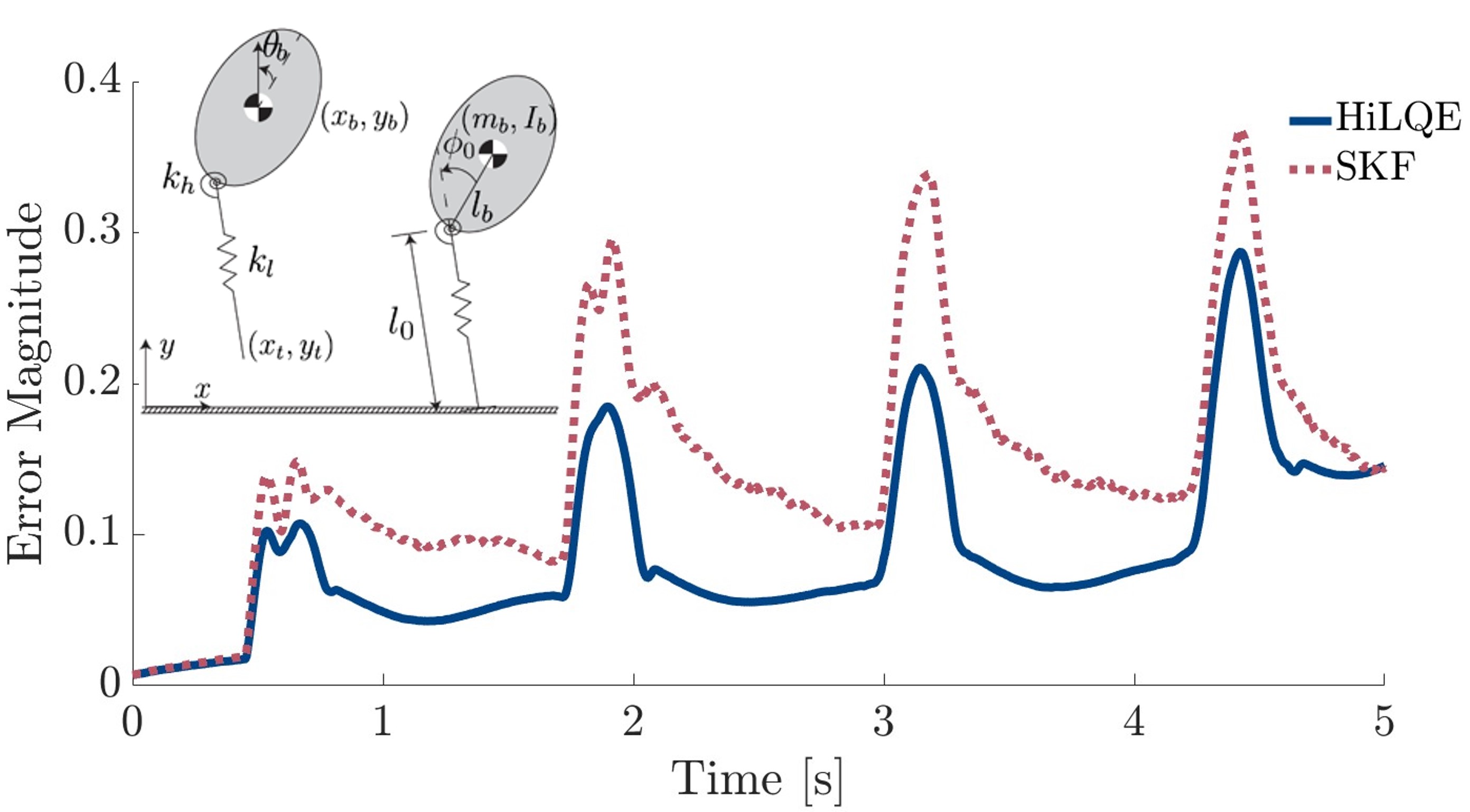}
    \caption{Average error magnitude over 100 trials when simulating a an ASLIP system (inset) which impacts the ground four times during its execution. The error from the proposed HiLQE method is shown as a solid blue line while the error from the SKF is shown as a dashed red line.
    }
    \label{fig:aslip_avg_error}
\end{figure}

To this end, here we propose Hybrid iterative Linear Quadratic Estimation (HiLQE), an offline iterative linear quadratic state estimation scheme for hybrid systems which optimizes over all sensor readings to achieve a more accurate state estimate. 
Similar to the Hybrid iterative Linear Quadratic Regulator (HiLQR) for control \cite{paper:kong-ilqr-2021,paper:kong-hybrid-2023}, we use the saltation matrix \cite{kong2024saltationmatricesessentialtool} in the backward pass of the iterative update in order to properly capture the gradients through impact events.
Additionally, we explicitly account for the differing hybrid system discrete modes \cite{Back_Guckenheimer_Myers_1993,LygerosJohansson2003,goebel2009hybrid, johnson2016hybrid}, utilizing the proper system dynamics for each mode during both the backward and forward passes in the smooth domains.
In the forward pass, we utilize a reference extension method \cite{paper:kong-ilqr-2021,paper:kong-hybrid-2023} to create reasonable comparisons between states in differing domains.
We demonstrate the efficacy of these methods on an asymmetric spring loaded inverted pendulum (ASLIP) system, which can be seen in Fig. \ref{fig:aslip_avg_error}.
With these updated techniques for hybrid transitions, HiLQE notably outperforms the online, incremental Salted Kalman Filter (SKF) method.

\section{Related Work}
As the fields of hybrid systems \cite{Back_Guckenheimer_Myers_1993,LygerosJohansson2003,goebel2009hybrid, johnson2016hybrid} and state estimation \cite{li2015kalman,cadena2016past} are relatively mature and have plenty of relevant work, here we focus specifically on work covering the intersection of the two.

\subsection{State Estimation for Hybrid Systems}
Initial works on Kalman filter (KF) based methods for hybrid dynamical systems such as \cite{bloesch2013state, hartley2020contact} utilize the reset map to update the mean estimate and the Jacobian of the reset map to update covariance beliefs at hybrid events.
However, recent work demonstrated that the Jacobian does not capture all of the effects of the hybrid event on reshaping the covariance. 
Instead, \cite{kong2021salt} proposed the Salted Kalman Filter, which uses the saltation matrix \cite{kong2024saltationmatricesessentialtool} in place of the Jacobian of the reset map at hybrid events for Kalman filtering. The saltation matrix includes a rank 1 update to the Jacobian that accounts for state variations caused by time-to-impact variations, and is discussed in more detail below.
This SKF will be the baseline against which we compare our accuracy in this work.
This work was extended to include filtering on manifolds in \cite{GAO2021290} with the Hybrid Invariant Extended Kalman Filter (HInEKF) and to include uncertainty in guard conditions and reset maps in \cite{paper:payne-uncertainty-2022} with the uncertainty aware Salted Kalman Filter (ua-SKF).
In this paper, the saltation matrix is used to determine gradient information in an optimization problem over a series of measurements.

Other work on hybrid system state estimation has largely involved multiple estimators in order to track belief in multiple hybrid modes. 
For example, the Interacting Multiple Model estimation algorithm (IMM) \cite{blom1988interacting} maintains KFs for each of the hybrid modes.
Multiple model methods have been extended to a variety of problems including nonlinear dynamics \cite{barhoumi2012observer} and non-identity rests \cite{BALLUCHI2013915}.
Multiple model methods are not easily applicable on event-driven hybrid dynamical systems as one of the core assumptions of these methods is that the transitions between discrete modes follow a Markov model, which is not necessarily true when the probabilities of discrete state transitions are dependent on the continuous state beliefs.

Alternatively, many methods such as \cite{koutsoukos2002monitoring,koval2017manifold} have adopted particle filtering approaches and use large numbers of individual estimates to represent a distribution as opposed to summary statistics like mean and covariance in the case of KFs. 
While these methods have many benefits, including capturing nonlinear dynamics and non-Gaussian beliefs, the computational complexity of running a particle filter is far greater than that of parametric filters which we seek to tune using our offline method.

In \cite{Zhang2020}, an offline optimization based state estimation method was proposed. This work relaxed the discrete state estimates to a continuous state in order to estimate both the continuous and discrete portions of the state simultaneously.
To handle the hybrid events in this continuous formulation, the model of the process noise was switched from a Gaussian distribution to a long tailed student's t distribution.
In contrast, the work we present in this paper explicitly considers the hybrid nature of the systems and accounts for non-identity reset maps directly.
By explicitly considering the dynamics in an iterative formulation, the estimates generated by our method will always be physically realizable, which could be useful in a potential extension to an online moving window approach.

\subsection{Nonlinear Event Mapping and Saltation Matrices}
\begin{figure}[tb]
    \centering
    \includegraphics[width = \linewidth]{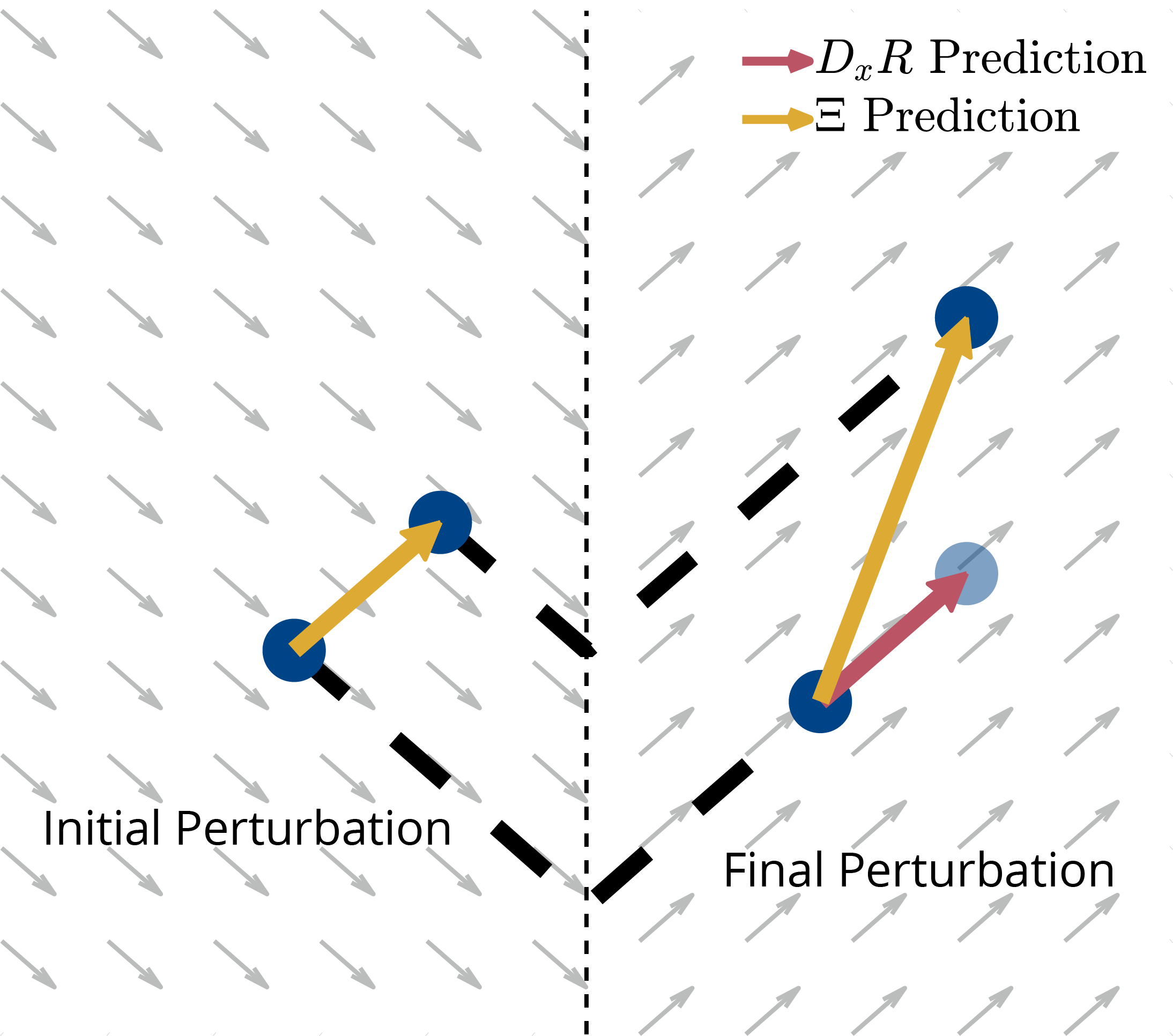}
    \caption{Evolution of a perturbation through a simple 2 mode hybrid system with constant dynamics in each mode. This demonstrates the importance of considering time-to-impact variations with the saltation matrix to obtain correct gradient information through hybrid transitions as the reset of this system is identity and the Jacobian of the reset map fails to capture these effects. First appeared in \cite{kong2024saltationmatricesessentialtool}.
    }
    \label{fig:gradient_information}
\end{figure}

The saltation matrix \cite{aizerman1958stability,hiskens2000trajectory,leine2013dynamics,burden2016event,kong2024saltationmatricesessentialtool} is used to map perturbations through non-smooth dynamics at transition events between discrete modes.
In contrast to the Jacobian of the reset map at a hybrid event, the saltation matrix captures to first order the effects of variations in dwell time in each hybrid mode caused by displacements in state.
The difference between the two methods of propagation through hybrid events can be seen in Fig. \ref{fig:gradient_information}.
In the context of the work we present here, the saltation matrix will be used to capture gradient information through hybrid events.

Previously, \cite{biggio2014accurate} demonstrated the saltation matrix can be used to map probability distributions through hybrid transitions in a quadratic formulation.
In this work we will use this to propagate quadratic cost functions backwards in an iterative linear quadratic estimation context.

\subsection{Iterative Linear Quadratic Methods for Hybrid Systems}
In recent work, Bailly presented the MAPE-DDP algorithm \cite{BaillyOptimalEstimation2021}, which uses a DDP/iLQ framework to estimate the centroidal state of a legged robot.
This is similar to prior work on smoothing filters such as the iterative extended Kalman smoother \cite{ypma2003iterativesmoothing} or the Backward Smoothing Extended Kalman Filter \cite{psiaki2005backward}.
A key difference between this approach for estimation and DDP or iLQR for path planning is that the Jacobian of the measurement function must be used in calculating cost gradients.
While a legged robot is a hybrid system, the floating base model presented does not treat the robot as a hybrid system. 
In contrast, the work we present in this paper explicitly considers the differing dynamics in each mode and the effects of mode transitions on gradient information.
By accounting for hybrid dynamics, our formulation would allow estimation to use information from lower level control architecture which deals with legged control directly. 

Another recent work solved the iLQR trajectory planning problem on hybrid systems \cite{paper:kong-ilqr-2021,paper:kong-hybrid-2023}. 
In that work the saltation matrix is used to calculate gradient information through mode transition events in the backward pass of an iLQR problem for trajectory planning.
The work we present in this paper can be viewed as the solution to the dual problem to the planning problem presented there.
One of the key differences is that the estimation is looking backward in time to estimate a trajectory whereas the planning problem looks forward to generate a trajectory.
Additionally, rather than a reference trajectory, the estimation problem we present in this work uses a series of past measurements, which may be in a lower dimensional space than the trajectory itself.

\section{Background}
\subsection{Hybrid Dynamical Systems}
A hybrid dynamical system is a system with continuous states, such as positions and velocities, and discrete states or modes, such as whether a specific limb is in contact with the ground. In a hybrid dynamical system, the sequence of discrete states is determined by the evolution of the continuous states. 
More formally, following \cite[Def.~1]{kong2024saltationmatricesessentialtool}:
\begin{definition} \label{def:hs}
    A $C^r$ \textbf{hybrid dynamical system}, for continuity class $r\in \mathbb{N}_{>0} \cup \{\infty,\omega \}$, is a tuple $\mathcal{H} := (\mathcal{J},{\mathnormal{\Gamma}},\mathcal{D},\mathcal{F},\mathcal{G},\mathcal{R})$ whose constituent parts are defined as:
    \begin{enumerate}
        \item $\mathcal{J} := \{I,J,...\} \subset \mathbb{N}$ is the set of discrete \textbf{modes}.
        \item $\mathnormal{\Gamma} \subset \mathcal{J}\times\mathcal{J}$ is the set of discrete \textbf{transitions} forming a directed graph structure over $\mathcal{J}$.
        \item $\mathcal{D}:=\amalg_{{I}\in\mathcal{J}}$ ${D}_{I}$ is the collection of \textbf{domains},
        \item $\mathcal{F}: \mathbb{R} \times \mathcal{D} \rightarrow \mathcal{TD}$ is a collection of $C^r$ time-varying \textbf{vector fields}, $F_I:=\mathcal{F}|_{D_I} :  \mathbb{R}\times D_I\to\mathcal{T}D_I$.
        \item $\mathcal{G}:=\amalg_{(I,J)\in\mathnormal{\Gamma}}$ $G_{I,J}(t)$ is the collection of \textbf{guard sets}, where $G_{I,J}(t)\subset D_I$ for each $(I,J)\in \mathnormal{\Gamma}$ is defined as a sublevel set of a $C^r$ function, i.e.\ $G_{I,J}(t)= \{x \in D_I|g_{I,J}(t,x)\leq0\}$.
        \item $\mathcal{R}:\mathbb{R}\times \mathcal{G}\rightarrow \mathcal{D}$ is a $C^r$ map called the \textbf{reset} that restricts as $R_{I,J}:=\mathcal{R}|_{G_{I,J}(t)}:G_{I,J}(t)\rightarrow D_J$ for each $(I,J)\in \mathnormal{\Gamma}$.
    \end{enumerate}
\end{definition}

\subsection{Perturbation Analysis and the Saltation Matrix}
Within a single mode, the Jacobian of the continuous dynamics can be used to obtain gradient information. 
Analogously, the saltation matrix is used to obtain gradient information through hybrid events.
The saltation matrix captures both the Jacobian of the reset map and the variations in time to impact as:

\begin{align}
\Xi_{(\mathrm{I},\mathrm{J})} := \der_x R^-+\frac{\left(F^+_{\mathrm{J}}-\der_xR^- F^-_{\mathrm{I}} - \der_tR^-\right)   \der_x g^-}{\der_t g^- +\der_x g^-  F^-_{\mathrm{I}}} 
\label{eq:saltation}
\end{align}
where $F^-_{\mathrm{I}}$ is the value of the first mode's vector field at the point of impact, $F^+_{\mathrm{J}}$ is the value of the second mode's vector field at the nominal reset point, $R^-$ is the reset map at the point of impact, and $g^-$ is the guard function representing the boundary between the two modes.
The saltation matrix captures both the effects of the Jacobian of the reset map and variations in the time a system is acted upon by the dynamics of each mode. It maps pre-transition variations $\delta x^-$ to post-transition variations $\delta x^+$ as, $\delta x^+ = \Xi \delta x^-$, as can be seen in Fig. \ref{fig:gradient_information}.

In order to ensure the saltation matrix is well defined, we utilize the conventional assumptions from \cite[Assumptions 1 and 2]{burden2018contraction}, which most notably include that the dynamics are transverse to guards and that Zeno behavior will not occur.
This ensures that trajectories in a neighborhood of a guard must transition exactly once at small timescales, which is a useful simplifying assumption when crafting algorithms to operate on these systems.

\subsection{Iterative Linear Quadratic Estimation}
\label{sec:iLQE}
The iterative Linear Quadratic Regulator (iLQR) algorithm \cite{todorov_ilqr} is traditionally used to solve trajectory optimization problems.
The same methods can be used to solve state estimation problems when properly formulated.
The state estimation optimization problem is written as \cite{BaillyOptimalEstimation2021}:
\begin{align}
\begin{split}
     \min_{X_{0:N},W_{0:N-1}} &  \frac{1}{2} \lvert\vert x_0-\bar{x}_0 \vert\rvert_{P_x}^2 \\
    & + \sum_{i=1}^{N}\left( \frac{1}{2} \lvert\vert h(x_i) -y_i \vert\rvert_{P_{v}}^2 + \frac{1}{2} \lvert\vert w_{i-1} \vert\rvert_{P_{w}}^2 \right) \\
    s.t. \quad   & x_{i+1} = f(x_i,u_i) + w_i \forall i
    \label{eq:optimizationproblem}
    \end{split}
\end{align}
Where $X$ and $W$ are the vectors of states and process noises, $x_i$ and $w_i$, for the full trajectory, $\bar{x}_0$ is the nominal initial state, $h(x)$ is a measurement function from the full state to the measurement space, $y_i$ is a measurement taken at timestep $i$, and $P_v$,$P_w$,$P_x$ are the inverses of the measurement noise, process noise, and arrival cost covariances respectively.
In this formulation, $\lvert\vert x \vert\rvert_{P}^2$ represents the squared weighted norm $x^TPx$.
This form of cost function is known as the Mahalanobis distance, which penalizes deviations in directions with more certainty more harshly than deviations in directions with low certainty.

Intuitively, this optimization simultaneously seeks to minimize process noise and deviations from measurements and the initial state while ensuring the dynamics are satisfied.

The iLQR algorithm seeks to solve this optimization problem by examining it in a series of backward-forward passes through a trajectory using the principle of optimality to separate the problem into a series of smaller optimal estimation problems.
This backward and forward iterative process is repeated until either some convergence property is met, such as the decrease in cost of a step falling below some threshold, or until a limit of iterations is reached.
\subsubsection{Backward Pass}

\begin{algorithm}[ht]
\caption{Backward Pass}\label{alg:backwardpass}
\begin{algorithmic}[1]
\State Initialize $X,W,Y$
\State $\V_{xx,N},\V_{x,N} \gets$ \Call{term\_grad}{$x_N,y_N$}
\For{$i \gets N-1:-1:1$}
    \If{i = 1}
    \State $l_x,l_{xx},l_{w},l_{ww} \gets$ \Call{init\_grad}{$x_i,w_i,y_i$}
    \Else
    \State $l_x,l_{xx},l_{w},l_{ww} \gets$ \Call{stage\_grad}{$x_i,w_i,y_i$}
    \EndIf
    \State $A,B_w \gets$ \Call{linearize\_dynamics}{$x_i,w_i$}
    \State $Q_x \gets l_x + A^T\V_{x,i+1}$ \Comment{\eqref{eq:Gradients}}
    \State $Q_w \gets l_w + B_w^T\V_{x,i+1}$
    \State $Q_{xx} \gets l_{xx} + A^T\V_{xx,i+1}A$
    \State $Q_{ww} \gets l_{ww} + B_w^T\V_{xx,i+1}B_w$
    \State $Q_{xw} \gets A^T\V_{xx,i+1}B_w$
    \State $k_i \gets Q_{ww}^{-1}Q_w$ \Comment{\eqref{eq:feedback_gain}}
    \State $K_i \gets Q_{ww}^{-1}Q_{xw}^T$
    \State $\V_{xx,i} \gets Q_{xx}-Q_{xw}Q_{ww}^{-1}Q_{xw}^T$ \Comment{\eqref{eq:opt_ctg}}
    \State $\V_{x,i} \gets Q_x - Q_{xw}Q_{ww}^{-1}Q_w$
    \State $\Delta J \gets \Delta J + Q_w^T k_i$ \Comment{Expected decrease}
\EndFor \\
\Return $K,k,\Delta J$
\end{algorithmic}
\end{algorithm}

Using the principle of optimality, the optimal cost-to-go $\V_k$ can be recursively written as:
\begin{align}
\begin{split}
    \V_i(x_i,u_i,w_i) = & \min_{w_i} l_i(x_i,u_i,w_i) \\
    & + \V_{i+1}(f(x_i,u_i,w_i))
\end{split}
\end{align}
Where $l_i$ is the cost associated with a single timestep in the optimization, which is represented by a single term in the summation in \eqref{eq:optimizationproblem}.
However, for ease of notation, we will suppress the $u_i$ in these expressions as the control input is assumed to be known and not optimized over.
The cost-to-go, $Q$, which will be optimized over to obtain $\V$ can be written as:
\begin{align}
    Q_i(x_i,w_i) = l_i(x_i,w_i) + \V_{i+1}(f(x_i,w_i))
\end{align}
This can be approximated with a second order Taylor expansion to determine cost change:
\begin{align}
    \Delta Q_i = & Q_i(x_i+\delta x_i, w_i + \delta w_i) - Q_i(x_i,w_i) \\
    \approx & \begin{bmatrix}
    1 \\ \delta x_i \\ \delta w_i
    \end{bmatrix}^T
    \begin{bmatrix}
    0 & Q_{x,i}^T & Q_{w,i}^T \\
    Q_{x,i} & Q_{xx,i}^T & Q_{xw,i}^T \\
    Q_{w,i} & Q_{xw,i} & Q_{ww,i}^T
    \end{bmatrix}
    \begin{bmatrix}
    1 \\ \delta x_i \\ \delta w_i
    \end{bmatrix}
    \label{eq:cost_change_expansion}
\end{align}
Where the block matrix entries can be written as:
\begin{subequations}
\label{eq:Gradients}
\begin{align}
    Q_{xx} = & l_{xx} + A^T\V_{xx,i+1}A \label{eq:Qxx}\\
    Q_{ww} = & l_{ww} + B_w^T\V_{xx,i+1}B_w \label{eq:Qww} \\
    Q_{xw} = & l_{xw} + A^T\V_{xx,i+1}B_w \label{eq:Qwx} \\
    Q_x = & l_x + A^T\V_{x,i+1} \label{eq:Qx}\\
    Q_w = & l_w + B_w^T\V_{x,i+1} \label{eq:Qw}
\end{align}
\end{subequations}
Where $B_w$ is the Jacobian of the dynamics function with respect to the process noise.
Using this second order Taylor expansion, the optimal update for $w_i$ can be calculated as:
\begin{align}
\label{eq:feedback_gain}
\begin{split}
    \delta w_i^* = & -Q_{ww,i}^{-1}Q_{xw,i}^T\delta x_i - \alpha_i Q_{ww,i}^{-1}Q_{w,i} \\
    \coloneqq & -K_i \delta x_i -\alpha_i k_i
\end{split}
\end{align} 
Where $\alpha_i$ is a line search parameter and $K$ and $k$ are defined to be the feedback and feedforward gains.

With the Taylor series approximation, the backward pass update to approximate the value function gradient and hessian at each timestep can be derived by substituting the optimal update into \eqref{eq:cost_change_expansion}:
\begin{subequations}
\label{eq:opt_ctg}
\begin{align}
    \V_{x,i} = & Q_{x,i} - Q_{xw,i}Q_{ww,i}^{-1}Q_{w,i} \\
    \V_{xx,i} = & Q_{xx,i} - Q_{xw,i}Q_{ww,i}^{-1}Q_{xw,i}^T
\end{align}    
\end{subequations}
These updates are then applied backward from the last measurement to the prior on the initial condition.
An outline of this process is found in Algorithm \ref{alg:backwardpass}.

\subsubsection{Initial Point Update}
Since the initial point is not known with certainty as in an iLQR control problem, an additional step must be added to update the initial point in the estimation problem \cite{BaillyOptimalEstimation2021}:
\begin{align}
    x_0' = x_0 - \V_{xx,0}^{-1}\V_{x,0}
\end{align}
This can be derived by looking at the approximation of the initial cost and taking the derivative of the expression with respect to a perturbation in x. The optimal point should have a derivative of zero:
\begin{subequations}
\begin{align}
    \V(x+\delta x) &= \V(x) + \V_x(x)\delta x + \delta x^T \V_{xx}(x)\delta x \\
    \frac{\partial}{\partial x}[\V(x+\delta x)] &= \V_x(x) + \V_{xx}(x)\delta x \\
    0 &= \V_x(x) + \V_{xx}(x)\delta x \\
    \delta x^* &= -\V_{xx}^{-1}\V_x
\end{align}
\end{subequations}
Practically, this is taken as a direction of improvement and is incremented by a search parameter $\alpha$ during a line search in the forward pass of the algorithm.

\subsubsection{Forward Pass}
After the backward pass determines the feedforward and feedback gains on the noise input, a forward pass is then performed to update the state and process noise estimates at each timestep.
The forward pass of the algorithm performs a line search with $\alpha$ over the feedforward gain $k$ calculated in the backward pass for each timestep with \eqref{eq:feedback_gain}.
The outline of this process is found in Algorithm \ref{alg:forwardpass}.

\begin{algorithm}[ht]
\caption{Forward Pass}\label{alg:forwardpass}
\begin{algorithmic}[1]
\State Initialize $X,W,J,K,k$
\State it $\gets 1$
\State $\alpha \gets 1$
\While{$J >  J_{goal} \And \text{it} < \text{max-itr}$}
    \State $X, W, J \gets$ \Call{rollout}{X,W,K,k,$\alpha$} \Comment{\eqref{eq:feedback_gain}}
    \State $\alpha \gets \alpha/2$ \Comment{Line Search}
    \State $\text{it}$++
\EndWhile \\
\Return $X,W,J$
\end{algorithmic}
\end{algorithm}



\section{Methods}
In this section we present the details of our proposed algorithm, Hybrid iterative Linear Quadratic Estimation (HiLQE).
As in the iLQE methods discussed in Sec.~\ref{sec:iLQE}, HiLQE will iteratively perform backward and forward passes until either a convergence parameter is met, such as a threshold in the cost decrease for a timestep, or until a set number of iterations is performed.

\subsection{Backward Pass}
The backward pass generates cost gradient information at each timestep so that the forward pass can update the process noise estimate to reduce the estimation cost.
In this work, we limit the class of hybrid systems to those with continuous measurement information through hybrid events.
This constraint ensures that the cost gradients are well defined near hybrid events.
In systems where the measurements are not continuous through hybrid events, there is mode information implicitly encoded in the measurements and a different approach should be taken to utilize that information.
In this pass, we deal with the smooth timesteps and the hybrid timesteps in different manners.
\subsubsection{Smooth Case}
In timesteps that did not experience hybrid events in the prior rollout, the cost gradients are calculated in the same way as in the standard estimation algorithm presented in Sec. \ref{sec:iLQE}. 
Based on the optimality principle, individual timesteps can be handled independently. 
Because of this, we do not need to account for hybrid events that occur in other timesteps when calculating gradients for a smooth timestep.
\subsubsection{Hybrid Case}
When the backward pass calculates the gradient information for a timestep that contains a hybrid event, the reset map and time-to-impact variations must be accounted for in the gradient information. 
To account for this, we augment the state and process noise Jacobians with the saltation matrix from \eqref{eq:saltation}. 
For simplicity, we assume that hybrid events occur at the end of timesteps. 
Using the saltation matrix we update the backward pass from \eqref{eq:Qxx}-\eqref{eq:Qw}:
\begin{subequations}
\begin{align}
    Q_{xx} = & l_{xx} + A^T\Xi^T\V_{xx,k+1}\Xi A \label{eq:QxxH}\\
    Q_{ww} = & l_{ww} + B_w^T\Xi^T\V_{xx,k+1}\Xi B_w \label{eq:QwwH} \\
    Q_{wx} = & l_{wx} + B_w^T\Xi^T\V_{xx,k+1}\Xi A \label{eq:QwxH} \\
    Q_x = & l_x + A^T\Xi^T\V_{x,k+1} \label{eq:QxH}\\
    Q_w = & l_w + B_w^T\Xi^T\V_{x,k+1} \label{eq:QwH}
    \end{align}
\end{subequations}

The saltation matrix $\Xi$ is calculated at the stored pre-impact state for each of the timesteps with hybrid events.

\subsection{Forward Pass}
The forward pass is used to update the state and noise estimates based on the information from the cost gradient of the previous backward pass. This step involves integrating the dynamics forward at each step based on updated noise inputs. As in the backward pass, the smooth and hybrid timesteps are handled differently.

\subsubsection{Smooth Case}
The dynamics that are used to integrate the state forward at each timestep are selected based on the currently estimated active mode as opposed to the standard algorithm where there is one set of dynamics applied for all timesteps.
In each timestep, the forward integration is augmented with an event function to check for situations where guard conditions for the currently active mode are met.
In cases where no guard condition is met, the feedforward and feedback gains are applied as they would be in a standard forward rollout step.

\subsubsection{Hybrid Case}
In contrast, when a timestep contains a hybrid transition, the forward pass needs to be modified.
In these cases, a guard condition will be met during a forward rollout.
At the instant when that guard condition is met, the reset map between the two hybrid modes will be applied and the dynamics will be updated from the first mode's dynamics to the second mode's dynamics.
From there, the integration will continue forward to the end of the discrete timestep under the new mode's dynamics.

The feedback gain calculation needs to be updated as well since reset maps may make trajectories that are very close together arbitrarily far apart in state space. 
An example of this is the case of a bouncing ball with a coefficient of restitution of 0.8 in which a the reference state has a velocity of $-10.1\frac{m}{s}$ slightly before impact and an estimated velocity for the current iteration of $+8\frac{m}{s}$ slightly after impact.
Nominally, the difference in velocity between these two states is $18.1\frac{m}{s}$.
However, in this case, these two states should be treated as very close together since the hybrid transition makes the post impact velocities very similar.
Without handling this, the estimate for the updated measurement noise can be updated far more than it should because of the large $\delta x$ that would be used in its calculation, as in algorithm \ref{alg:forwardpass}.

This must be handled by reference extension for feedback gain calculation. 
In the case of reference extension, the current estimated state should not be directly compared to the state estimate from the prior forward pass.
Instead it will be compared to a value meant to reflect the prior estimate's distance from the current estimate projected into the current mode.

In cases where the new rollout has not made a transition and the reference state has, we back up to the impact point of the reference state and integrate it forward with the original dynamics to obtain the reference state.
In cases where the new rollout has made a transition before the former trajectory, we combine the difference between the current state and the nominal post-impact state from the prior rollout with the difference between the current reference state and the nominal reference impact state propagated with the saltation matrix to map that displacement into the new mode:
\begin{align}
    \delta x = X_n(i) - r(X_{ref}(j)) + \Xi (X_{ref}(j)-X_{ref}(i))
\end{align}
Where $X_n$ is the current rollout, $X_{ref}$ is the reference state from the prior forward pass, $i$ is the current timestep of interest, and $j$ is the timestep at which the reference state reaches the hybrid event.

\section{HiLQE Performance on Sample Systems}
\subsection{Bouncing Ball}
In this section, we present the estimation performance on a two-dimensional bouncing ball system.
The state of the system is represented as:
\begin{align}
    q = [x,y,\dot{x},\dot{y}]^T
\end{align}
Additionally, the coefficient of restitution of the system was $0.8$.
For this test, the position of the center of mass is measured, and not the velocity to ensure that the measurements are continuous near the impact event.
The data for this analysis was collected over 1000 one second long trials with measurements at 100Hz, which results in a single impact event during each trial.
The initial conditions for the trial were $x = 0, y = 1, \dot{x} = 0.5, \dot{y} = -5$.
The system was corrupted with a process noise of covariance $0.1I_4$ and the measurement was corrupted with covariance $I_2$.

\begin{figure}[tb]
    \includegraphics[width = \linewidth]{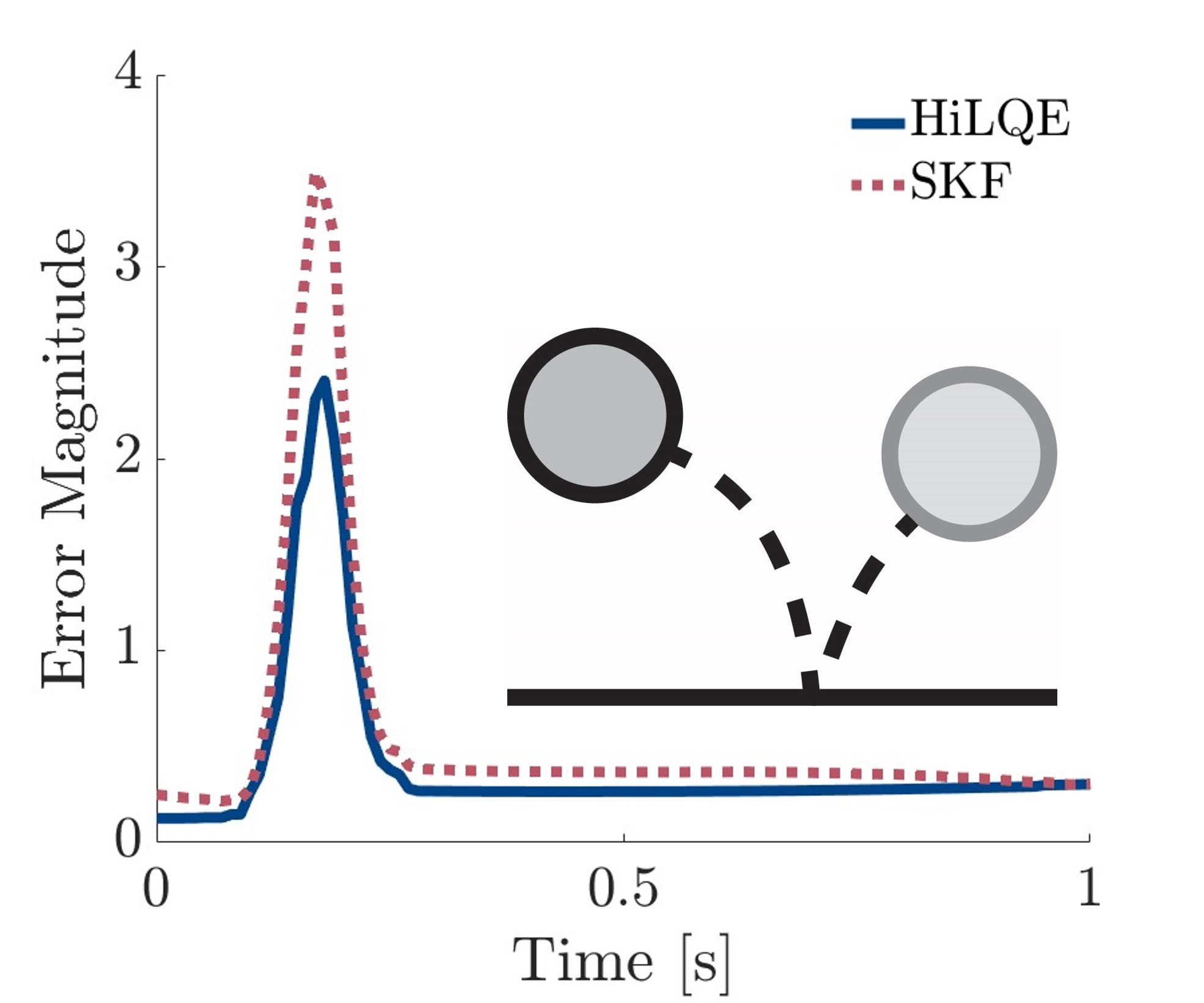}
    \caption{Average error magnitude over 1000 trials when simulating a bouncing ball system (inset) which impacts the ground once during its execution. 
    The errors from the proposed method are shown in blue solid lines while the errors from the salted Kalman filter are shown in dashed red lines.
    }
    \label{fig:bouncing_ball_error}
\end{figure}

The comparison of the total magnitude of the error as well as an illustration of the system can be seen in Fig.~\ref{fig:bouncing_ball_error}.
The HiLQE method outperformed the SKF at a peak of 48.60\% for a timestep near the impact event, and the median MSE trajectory improvement across all of the trials was 30.48\%.

This improvement arises out of HiLQE's ability to shift around the impact timing with more measurement information.
Associated with this, HiLQE achieves 9.8\% better mode estimation accuracy around impact events than SKF.

\begin{figure*}[tb]
    \centering
    \includegraphics[width = \textwidth]{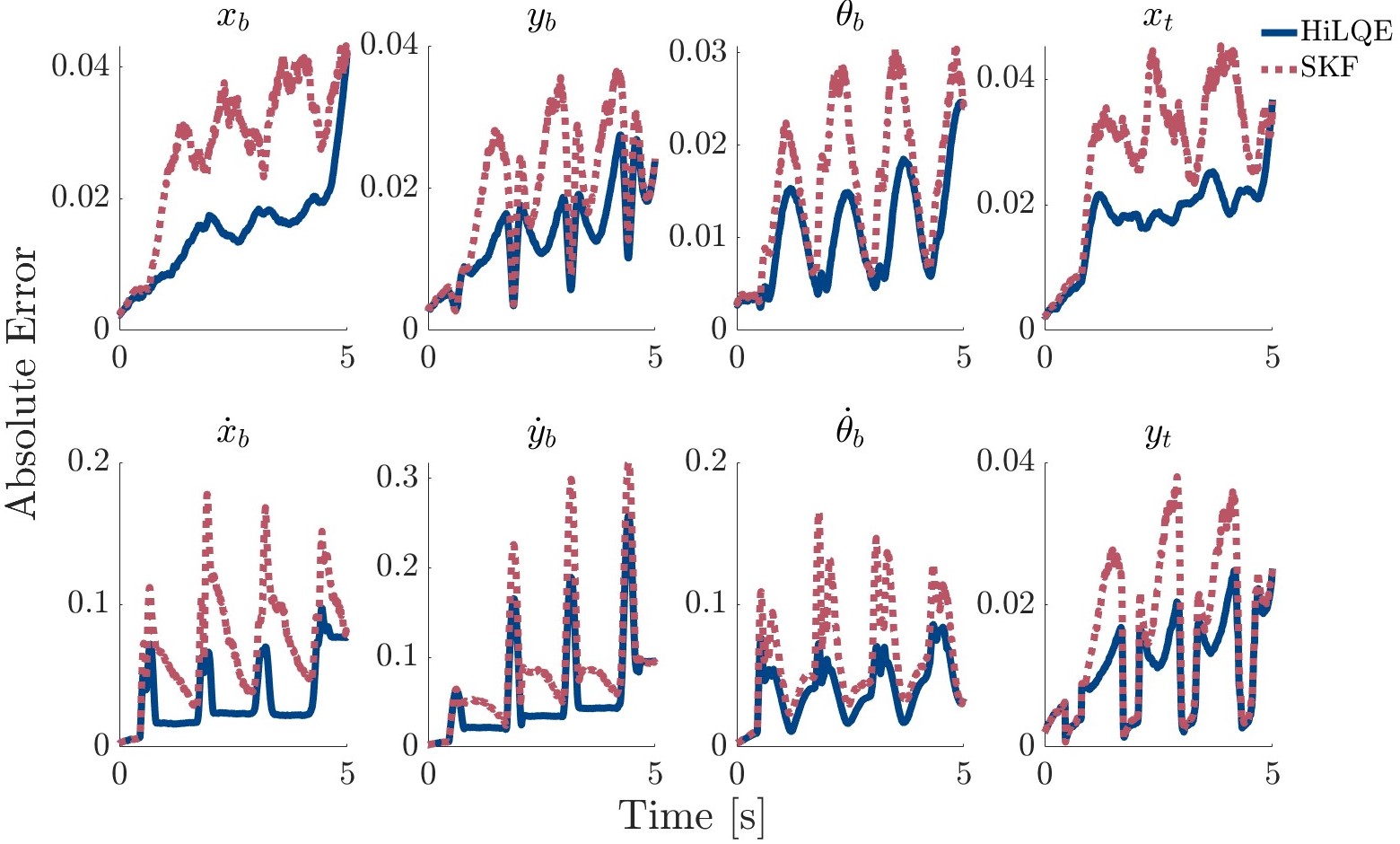}
    \caption{Average error for each state over 100 trials when simulating an ASLIP system which impacts the ground four times during its execution. The errors from the proposed method are shown in blue solid lines while the errors from the salted Kalman filter are shown in dashed red lines.
    }
    \label{fig:aslip_avg_error_all_dims}
\end{figure*}

\subsection{ASLIP Hopper System}
We demonstrate the efficacy of this method on the ASLIP (asymmetric spring loaded inverted pendulum) hopper \cite{poulakakis2009slip}, which can be seen in Fig.~\ref{fig:aslip_avg_error}.
The state for the ASLIP hopper is defined as:
\begin{align}
    q = [x_b,y_b,\theta_b,x_t,y_t,\dot{x}_b,\dot{y}_b,\dot{\theta}_b]^T
\end{align}
In the aerial phase, the system uses ballistic dynamics with a fixed leg length and angle.
We used Lagrangian dynamics to generate the dynamic equations for the contact mode.
The guard for touchdown when $y_t = 0$, or when the toe height reaches zero.
The guard for liftoff is $l_l -l_0 = 0$, where $l_l$ is calculated from the displacement between the toe and the hip attachment point to the body and $l_0$ is the rest length of the leg spring. The full derivation of the dynamics is presented in \cite{kong2021salt}.

For the results presented in this work, the data is gathered from 100 five second long trials at a sample rate of 1 kHz, which typically results in four jumps, depending on the sampled noise.
The parameters used for these trials were:
\begin{align}
\begin{split}
m_b = 1\mathrm{kg}, a_g = 9.8\mathrm{m/s^2}, l_b = 0.5\mathrm{m}, I_b = 1\mathrm{kg m^2}, \\
\nonumber k_h = 100\mathrm{nm/rad}, k_l = 100\mathrm{n/m}, l_0 = 1\mathrm{m}, \phi_0 = 0\mathrm{rad}
\end{split}
\end{align}

For these trials, the state was nominally initialized at $y_b = 2.5\mathrm{m}$, $y_t = 1\mathrm{m}$, with all other states and velocities set to zero.
This initial state is corrupted with a covariance of $10^{-5}I_8$.
The process noise sampled for these trials was $W = 10^{-1}I_8$ and the measurements were taken for all positions (but not velocities) with measurement covariance $V = I_5$.

The comparison of the total magnitude of the error across all dimensions can be seen in Fig.~\ref{fig:aslip_avg_error}, and a breakdown of the error reduction in each dimension of the state can be seen in Fig.~\ref{fig:aslip_avg_error_all_dims}.
The peak average percent improvement for a single timestep was 63.55\% throughout these trials, which indicates that in many of the spikes near impact, the HiLQE algorithm strongly outperforms the SKF.
Similarly, the median MSE improvement throughout entire runs was 61.96\% throughout the 100 trials as HiLQE consistently outperforms the SKF at all timesteps within these trials.

\section{Conclusion and Future Work}
In this paper we presented HiLQE, an offline method for estimating the state of hybrid systems. 
We used the saltation matrix to obtain gradient information through hybrid events to approximate the value function gradient and hessian at each timestep in the backward pass.
In the forward pass we modified the dynamics to include checking for guard conditions and applying reset maps.
Additionally, when calculating feedback gains in the forward pass, we applied a reference extension method to deal with mode mismatch due to shifting estimates of impact times.

The performance of the hybrid estimator was demonstrated on an ASLIP hopper model. This algorithm produced notably better results than the comparable SKF.
In the presented trials, HiLQE outperformed SKF by a peak of 63.55\% at the worst timesteps near impact events.

There are still a wide variety of interesting research directions in hybrid state estimation.
An immediate extension to this work would be to implement this iterative method online over a moving horizon of state estimates in an analogous manner to Model Predictive Control (MPC) for control problems.
This work could also be combined with information seeking control.
To deal with uncertainty in hybrid mode, we could look to designing control schemes that seek to gain as much confidence in the current mode as possible while still achieving other control goals.
In legged systems, an optimization like this might result in a behavior such as slamming legs into the ground rapidly to ensure contact is made.
Another potential direction for future work is platform and controller design for hybrid systems.
One interesting aspect of this would be to investigate whether it is possible to optimize robot designs and gaits in a way that the dynamics can be relatively accurately expressed as smooth systems.
This would save on the need to consider time-to-impact variations and simplify the gradient computations.
Additionally, this work could be integrated with the Simultaneous Localization And Mapping (SLAM) problem by treating the solutions to the visual estimations as measurements on the system state in this optimization framework.



\addtolength{\textheight}{-12cm}
\bibliographystyle{IEEEtran}
\bibliography{references}
\end{document}